\documentclass[lettersize,journal]{IEEEtran}
\usepackage{amsmath,amsfonts}
\usepackage{algorithmic}
\usepackage{algorithm}
\usepackage{array}
\usepackage[caption=false,font=normalsize,labelfont=sf,textfont=sf]{subfig}
\usepackage{textcomp}
\usepackage{stfloats}
\usepackage{url}
\usepackage{verbatim}
\usepackage{graphicx}
\usepackage{cite}
\usepackage{xcolor}
\begin{document}

\title{UFed-GAN: A Secure Federated Learning 
Framework with Constrained Computation and Unlabeled Data}

\author{Achintha Wijesinghe, Songyang Zhang,~\IEEEmembership{Member,~IEEE}, Siyu Qi and Zhi Ding~\IEEEmembership{Fellow,~IEEE,} 
\thanks{A. Wijesinghe, S. Qi, and Z. Ding are with the Department of Electrical and Computer Engineering, University of California, Davis, CA, 95616.  (E-mail: achwijesinghe@ucdavis.edu, syqi@ucdavis.edu, and zding@ucdavis.edu).

S. Zhang was with the University of California at Davis, Davis, CA, 95616 and now is with the Department of Electrical and Computer Engineering, University of Louisiana at Lafayette, Lafayette, LA 70504  (E-mail: songyang.zhang@louisiana.edu).}
\vspace*{-6mm}
}



\maketitle

\begin{abstract}
To satisfy the broad applications and insatiable hunger for deploying
low latency multimedia data classification and data privacy in a cloud-based setting, federated learning (FL) has emerged as an important learning paradigm.
For the practical cases involving
limited computational power and only unlabeled data in many wireless communications applications, this work investigates FL paradigm in a resource-constrained and label-missing environment. Specifically, we propose a novel framework of UFed-GAN: Unsupervised Federated Generative Adversarial Network, which can capture user-side data distribution without local 
classification training. We also analyze the convergence and privacy of the proposed UFed-GAN. Our experimental results demonstrate the strong potential of UFed-GAN in addressing limited computational resources and unlabeled data while preserving privacy.
\end{abstract}


\section{Introduction}
\IEEEPARstart{T}{he} burgeoning rise of deep learning has shown remarkable achievements in learning, based on the often voluminous amounts of data for centralized training. However, in many cases of
learning-based wireless connections
for collaboration, decentralized learning is vital to handle the heterogeneous data distribution among nodes (users). Importantly, privacy concerns and resource limitations also prevent direct data sharing. To ensure data privacy and communication efficiency, federated learning (FL)~\cite{fedAvg} has emerged as an important framework to disengage data collection and model training via local computation and global model aggregation.

Despite reported successes, existing FL frameworks have certain limitations. One major obstacle of 
FL in practice is the heterogeneity of data distribution among participating FL users. It is known
~\cite{non-iid1} that the accuracy of classic FL frameworks such as
FedAvg~\cite{fedAvg} could drop by 55\% for some datasets showing
non-IID, i.e., not identically and independently distributed, data distributions. To combat performance loss against non-IID datasets, the
more general approach of FedProx~\cite{fedprox}
may depend on certain unrealistic dissimilarity assumptions of local functions \cite{yuan2022convergence}. Alternatively,  generative adversarial network (GAN)~\cite{GAN1} provides another approach to address data heterogeneity. In GAN-based FL, GAN models are used as a proxy to share user updates without training the global model on the user side. For example, in~\cite{cGAN1}, a global classifier is trained using a user-shared generator of the user-end-trained conditional GANs (cGANs). Another example is~\cite{fullGAN1}, the authors proposed to share the full user-end GAN for generating a synthetic dataset. However, the high communication cost and the potential privacy leakage hinder the performance gain of these 
known GAN-based FL frameworks as discussed in~\cite{our}. Moreover, the training of the entire GAN and other learning models sometimes can be impractical at the user end, especially for nodes with limited computation resources, such as computation-constrained sensors and devices \cite{mu2020computation}. How to develop a more efficient GAN-sharing strategy to preserve privacy and handle limited computation remains an open question in generative FL.

In addition to data heterogeneity and limited computation resources, most of the existing works focus on supervised FL, where the performance depends heavily on the availability of labeled training data. In practical applications such as user clustering and video segmentation \cite{learn_from_unl}, the computational and privacy limitations may prevent user-side data labeling. Therefore, learning from unlabeled data is equally important for FL to reach its full potential. 
Presently, only limited research 
works have specifically addressed FL with unlabeled data, primarily due to inherent challenges. A typical category of FL dealing with unlabeled data~\cite{cluster2,cluster1} focuses on clustering tasks, which may limit its generalization to other deep learning tasks. Another line of FL focuses on unsupervised representation learning \cite{han2022fedx}, where knowledge distillation and contrast learning are applied to address heterogeneous user data distributions. 
Other FL works on unlabeled data~\cite{9348203,unlabeledFL,orchestra} leverage the efficiency of latent space and may lack a general description of the original data. As aforementioned, GAN-based approaches can be intuitive solutions to capture the data distributions and assist further applications, even without annotated labels.

In this work, we develop a novel FL framework, the Unsupervised Federated Generative Adversarial Network (UFed-GAN), for resource-limited distributed users without labeled data. The novelty is an innovative GAN-based FL and data-sharing strategy 
to significantly reduce the computational cost at the user end and 
to preserve privacy. Note that, instead of focusing on one specific unsupervised learning task, we provide a general FL scheme to learn the data distributions without labels. Our framework can be easily adapted to
handle specific learning tasks, including unsupervised representation learning, user clustering, and semi-supervised classification.

Our contributions can be summarized as follows:
\begin{itemize}
    \item We propose a novel UFed-GAN as an FL framework to learn and
    characterize the non-IID user data distributions from unlabeled user data. Our UFed-GAN captures the underlying user data distributions without explicitly training a local GAN model for each user, thereby significantly lowering the computational cost on the user side. To our best knowledge, this is the first work to address such constrained computation in GAN-based FL.
    \item We analyze the convergence of UFed-GAN and prove that privacy leakage can be prevented by our UFed-GAN, in comparison to traditional GAN-based FL.
    \item Our experimental results in several benchmark datasets demonstrate the performance of UFed-GAN in a semi-supervised classification setup.
\end{itemize}

In terms of organization, we first introduce the architecture of UFed-GAN and a training strategy in Section~\ref{sec:2}. Following the study on model convergence and 
the privacy analysis in Section ~\ref{sec:3}. We present the experimental results of UFed-GAN on several well-known datasets in Section~\ref{sec:4}. We
provide concluding remarks in Section~\ref{sec:5}.

\section{Method and Architecture}
\label{sec:2}
\subsection{Problem Setup}
\label{ss:ps}

\begin{figure}[!t]
\centering
\includegraphics[width=3in, height= 1.3in]{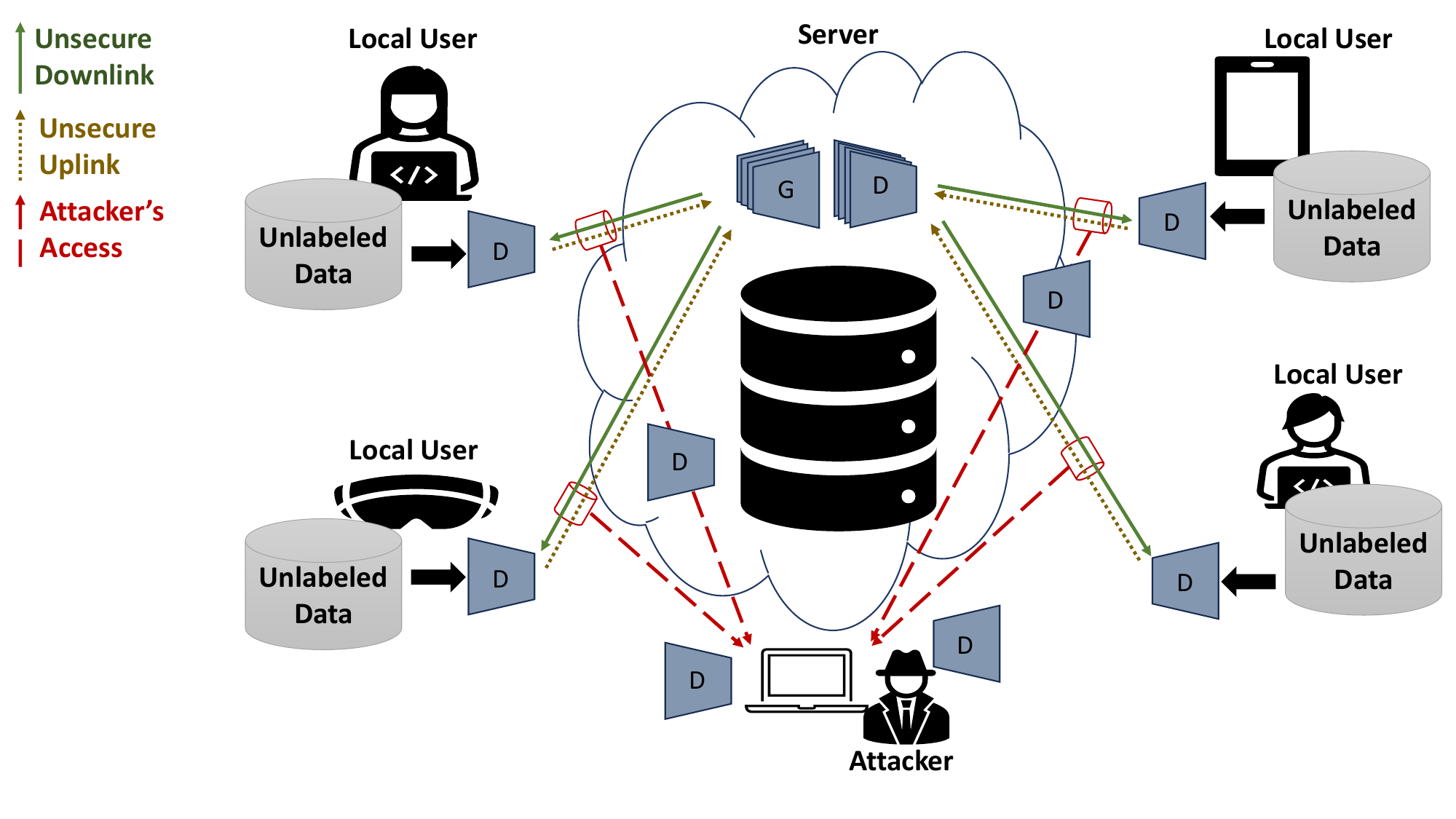}
\vspace{-2mm}
\caption{UFed-GAN in a distributed learning setup in an untrustworthy communication scenario, where users are assumed with less computational power. An attacker may eavesdrop to understand the user data.}
\vspace{-3mm}
\label{fig_1}
\end{figure}

As a typical example of a resource-limited FL setup in Fig.~\ref{fig_1}, a server aims to learn from 
user nodes, each with limited computational resources whereas attackers may attempt to eavesdrop on the network links. Users may not be able to annotate their raw data. Moreover, since the target tasks may be different among users and the data may also be skewed, a non-IID data distribution shall be considered in this scenario. Different from local users, the server has sufficient computational resources to obtain a model that learns a global data distribution from all the local data, without initial training data. Such a setup is applicable in many distributed learning scenarios. For example, a distributed camera/sensor system placed for object detection can benefit from the collaboration of different cameras for better feature extraction, where each digital
camera or sensor may have limited computation power. 

To demonstrate the privacy protection offered by UFed-GAN, we consider an attacker that has access to the vulnerable communication links between distributed users and the central server. Such attacks try to gain users' data features based on the information shared through the channels.

Note that, we aim to develop a novel data/model-sharing strategy for FL. The strategy could handle unlabeled data and capture user data distributions in the scenario with limited local computation resources. The proposed framework should offer flexible integration with various unsupervised and 
semi-supervised learning tasks, such as latent representation learning, user clustering, and semi-supervised classification~\cite{han2022fedx}.

\subsection{UFed-GAN}
\label{ss:uf}

\begin{figure}[!t]
\centering
\includegraphics[width=3.2in, height= 1.3in]{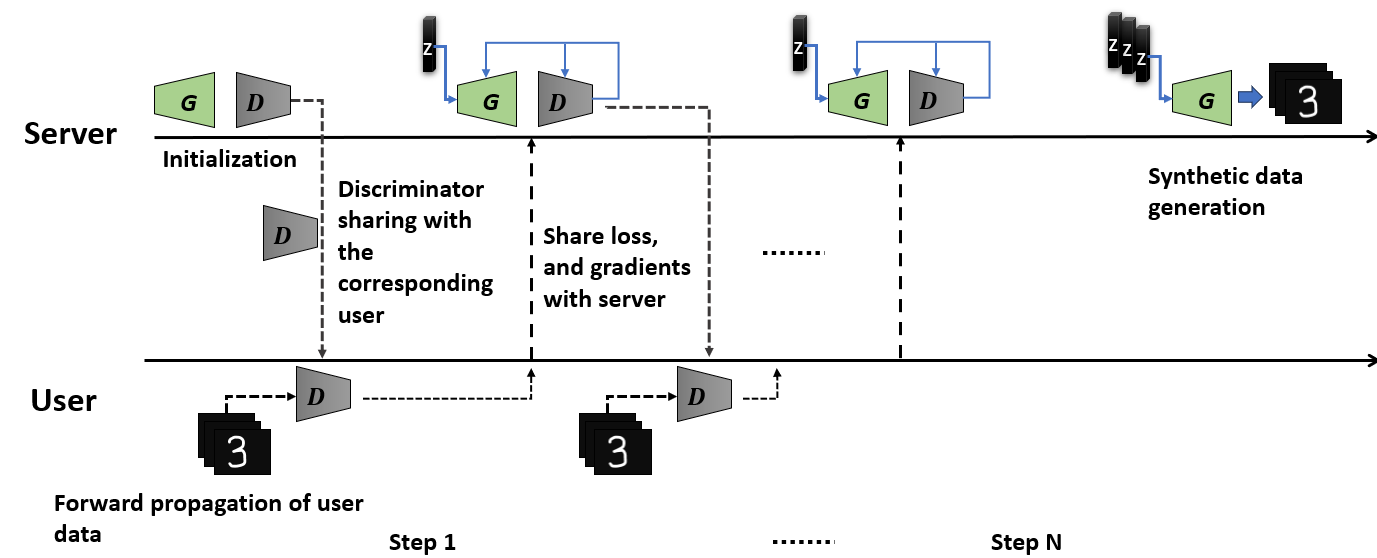}
\caption{ Communication rounds till the convergence of UFed-GAN. We use the inception score (IS) as the measure of convergence. }
\vspace{-3mm}
\label{fig_com}
\end{figure}

We now explain the framework and training process of UFed-GAN, which allows private and secure learning from unlabeled data in a distributed learning setup. Our proposed UFed-GAN aims to train a GAN model on the server to capture the underlying user data distributions without implementing complex GAN training on the user side.

First, for each user $u$, we initiate a GAN model on the server, including a generator ($G_u$) and a discriminator ($D_u$). Due to label inaccessibility, we select deep convolutional GANs (DCGANs)~\cite{dcGAN} as the backbone. The details of choices for GAN architecture will be elaborated in Section~\ref{arc}.

The GAN training comes in three steps, which contains two steps of $D$ training and one step of $G$ training.
We split the dual-step $D$ training into the server side and the user side. On one side, the server initiates a discriminator $D_u$ and then shares the initiated model with the corresponding user $u$. 
Subsequently, on the other end, the user performs a forward pass (FP) on $D_u$ using a single batch of real local data $I_u$. The gradients and loss are calculated and then shared with the server. Compared with training a full GAN on the user side, this step can significantly reduce the computational cost and be easily deployed in computation-constrained devices.
Upon receiving the updated information of $D_u$ from the user $u$, the server completes the training of $D_u$. It generates a noise vector $\textbf{z}$ to pass through $G_u$. Finally, synthetic data $G_u(\textbf{z})$ is generated from the generator.
$G_u(\textbf{z})$ is then sent to $D_u$ to calculate the corresponding gradients and loss. These received gradient
updates are combined with the gradient updates of the previous step and then back-propagated through $D_u$ to update the parameters. 

Similar to conventional GAN training, starting with a noise vector $\textbf{z'}$, we train $G_u$ with forward- and backward-propagation to update its parameters. In each communication round, the above process repeats until GAN convergence to a favorable point. We illustrate this training process further in Fig.~\ref{fig_com}.
For model convergence monitoring and the stopping criterion, we use inception score (IS)~\cite{IS}, which measures the characteristics of the generated images. After convergence, we create a synthetic dataset using the trained $D_u$. With the generated synthetic dataset, we can design corresponding unsupervised or semi-supervised algorithms to implement specific learning tasks. For example, in a semi-supervised classification task, we could apply the MoCo~\cite{moco}, which uses dictionary lookups in contrastive learning to obtain the global classifier. The pseudocode of the proposed training strategy is presented in Algorithm~\ref{alg:alg1}.

\begin{algorithm}[h]
\caption{UFed-GAN: Training Algorithm}\label{alg:alg1}
\begin{algorithmic}
\FOR{ each user $u$}
\STATE Server initialization of $G_u$ and $D_u$ 
\ENDFOR
\FOR{ each communication round $r$, until the GAN convergence}
\FOR{ each step $t$}
\STATE Share $D_u$ with user $u$.
\STATE Perform FP in $D_u$ with user data $\mathcal{I}_u$ and share the gradient updates $\nabla W_u$. 
\STATE Perform FP in $D_u$ with fake data $G(z_t)$ where $z_t$ is a random noise vector and get the gradient updates $\nabla W_z$.
\STATE Update $D_u$ with ($\nabla W_u + \nabla W_z$).
\STATE Train $G_u$ with trained $D_u$ and random noise vectors.
\ENDFOR
\STATE Generate an unlabeled dataset using each $G_u$.
\ENDFOR
\end{algorithmic}
\label{alg1}
\end{algorithm}

\subsection{Attacker Model} \label{atk}
We now introduce the attacker model to quantify the privacy leakage of UFed-GAN. It is highly challenging, if not impossible, to obtain access to the global generator $G$ given a secure server or to guess the exact architecture of $G$, regardless of the computation prowess of the attacker. Therefore, as suggested in~\cite{our}, we focus on a reconstruction attack~\cite{hayes2017logan} in this work, where an attacker attempts to reconstruct the training data.
Let $\theta$ be the parameters released to the communication channel by the user. We denote the releasing mechanism by $\mathcal{M}$ and the information that an attacker $\mathcal{A}$ could obtain by $\mathcal{M}(\theta)$. 
According to the UFed-GAN framework, $\mathcal{M}(\theta)$ is the discriminator gradients and the loss values. Let $\mathcal{I}$ represent the reconstructed data, we have
\begin{equation}
    \mathcal{A}: \mathcal{M}(\theta) \mapsto \mathcal{I}.
\end{equation}

Suppose that the generator $G_{\mathcal{A}}$ of attacker $\mathcal{A}$ has the same architecture as $G$ but with initial weights $W_{\mathcal{A}}$ different from those of $G$, represented by $W_G$. In parallel to the server training, we train $G_{\mathcal{A}}$ at the attacker's end.

\subsection{GAN Architecture} \label{arc}


Due to label inaccessibility and resource constraints, we adopt the unsupervised DCGANs~\cite{dcGAN} over cGANs \cite{our} which saves extra computation needs for any pseudo-labeling.
The generator architecture follows five transposed convolutional layers. The first layer takes an input with 100 channels and maps it to 1024 channels. Every subsequent layer reduces the number of channels by half. 
Every layer 
uses ReLU activation except 
for the Tanh activation for the final layer. All the layers in both the generator and the discriminator use $4\times4$ kernels and batch normalization for each layer before the final layer. For the discriminator model, we use four convolutional layers. The first layer accepts similar channel sizes of the data samples and maps to 256 channels. Every subsequent layer doubles the number of channels except the final layer, which outputs a single channel. The final layer uses a Sigmoid activation whereas all other layers use LeakyReLU activation.

\section{Convergece and Privacy Analysis of UFed-GAN}
\label{sec:3}
In this section, we present the convergence and privacy analysis of the proposed UFed-GAN. We introduce the major proof steps and refer those interested to our corresponding references for more details.

\subsection{Convergence of the discriminator}

Let $\mathcal{G}$ and $\mathcal{D}$ be the generator and the discriminator of a GAN, respectively. Assume $\mathcal{G}$ is capable of capturing a distribution $p_{GS}$ on the server and we are interested in learning a user distribution $p_{data}(x)$.\\

\noindent {\textbf{Proposition 1.}}
{\em Any $\mathcal{D}$ initiated on a server and trained in accordance to Algorithm~\ref{alg:alg1} with $\mathcal{G}$ and $p_{data}(x)$ converges to a unique $\mathcal{D}^{*}$ for the given $\mathcal{G}$ as presented in~\cite{GAN1}, i.e.,
\begin{equation}
    \mathcal{D}^{*} = \frac{p_{data(x)}}{p_{data(x)}+p_{GS}}
\end{equation}
}

Since UFed-GAN merely splits the GAN training, the proof of Proposition 1 shall directly follow that in \cite{GAN1}.
This proposition serves as a guarantee of the convergence of the server-side discriminator. It shows that the discriminator is still capable of capturing the user side's data distribution. This helps the generator on the server-side to generate user-like data as illustrated in the following proposition.

\subsection{Convergence of the generator}
\noindent{\textbf{Proposition 2.}} 
{\em Any $\mathcal{G}$ initiated on a server and trained in accordance to Algorithm~\ref{alg:alg1} with $\mathcal{D}$ and $p_{data}(x)$ converges to a unique $\mathcal{G}^{*}$ which captures $p_{data}(x)$. i.e. $p_{GS} = p_{data}(x)$.}\\

Since UFed-GAN does not alter the training of $\mathcal{G}$, we can imitate and adopt the proof steps in~\cite{GAN1}. Proposition 2 suggests that the server-side generator converges to the same generator that
could have been trained locally. Therefore, on the server, we are able to regenerate synthetic samples which resemble the user data in terms of data distribution.

\subsection{Divergence of any discriminator other than $\mathcal{G}$}

\noindent{\textbf{Proposition 3.}} 
{\em Any generator $G$, other than  $\mathcal{G}$ trained in accordance to the Algorithm~\ref{alg:alg1} with $\mathcal{D}$  diverges from unique $\mathcal{G}^{*}$. i.e. $p_{GS} \neq p_{data}(x)$.}\\

To prove Proposition 3, we adopt a similar proof process as provided in 
\cite{our}. Following Proposition 1 and Proposition 2, the pair of $\mathcal{G}$ and $\mathcal{D}$ is unique. Therefore, at each training step $\mathcal{G}^{'}=G^{'}$ must be asserted. Hence, any $G$ difference from $\mathcal{G}$ at any step fails to capture $p_{data}(x)$.

\section{Results and Discussion}
\label{sec:4}
In this section, we present the experimental results on both utility and privacy.
\subsection{Evaluation of the Utility}
In a common semi-supervised setting as~\cite{han2022fedx}, let $N$ be the number of users, $\beta$ be the concentration parameter of Dirichlet distribution ($Dir_N(\beta)$), and $s_{ij}$ be a sample taken from $Dir_N(\beta)$. We assign $s_{ij}$ in proportion to the $i$-th class size of the user $j$. We pick $N=10$ and $\beta=0.5$ in accordance with~\cite{han2022fedx}. All model comparisons are based on the linear evaluation protocol, which trains a linear classifier on top of representations or regenerated fake data \cite{zhang2020federated}.

\begin{table}[t]
\label{tab:fid}
\caption{Classification accuracy comparison of different FL approaches over three datasets. Part of the results in this table are reported from~\cite{han2022fedx}.}
\centering
\begin{tabular}{|c||c||c||c|}
\hline
Method & CIFAR 10 & SVHN & FashionMNIST\\
\hline
FedSimCLR  & 52.88 & 76.50 & 79.44\\
+ FedX     & 57.95 & 77.70 & 82.47\\
\hline
FedMoCo  & 57.82 & 70.99 & 83.58\\
+ FedX & 59.43 & 73.92 &  84.65 \\
\hline
FedBYOL  & 53.14 & 67.32 & 82.37\\
+ FedX  & 57.79 & 69.05 & 84.30\\
\hline
FedProtoCL & 52.12 & 50.19 & 83.57\\
+ FedX & 56.76 & 69.75 & 83.34\\
\hline
FedU  &  50.79 & 66.22 & 82.03\\
+ FedX & 57.26 & 68.39 & 84.12 \\
\hline
Full GAN & 68.77 & 80.17 & 86.25\\
\hline
\textbf{UFed-GAN} & 67.0 & 80.109 & 86.33 \\

\hline
\end{tabular}
\vspace{-2mm}
\end{table}

We consider three well-known datasets: CIFAR10~\cite{cifar10}, SVHN~\cite{svhn} and FashionMNIST~\cite{fashionMnist}. Comparative results against five other FL algorithms are 
presented in Table~\ref{tab:fid}.
These algorithms are: FedSimCLR \cite{chen2020simple}, FedMoco \cite{moco}, FedBYOL \cite{grill2020bootstrap}, FedProtoCL \cite{li2020prototypical} and FedU \cite{zhuang2021collaborative}. For each method, we present their accuracy on the respective dataset, together with their accuracy when further applying FedX \cite{han2022fedx}. We also compare the results with ``full GAN'' sharing. 

From the results, UFed-GAN outperforms all other FL methods in the benchmark group, with an improvement of around 8\% in CIFAR10, 3\% in SVHN, and 2\% in FashionMNIST. 
The accuracy gain arises from the power of GANs to understand the underlying data distribution of users and to generate synthetic data by preserving essential features. 
Another observation is that UFed-GAN is 
dataset-agnostic in terms of performance, delivering the best outputs
in all tested datasets.
This observation promotes the generalizability of the proposed method. 
In fact, UFed-GAN achieves similar performance as with full GAN sharing. 
However, full GAN sharing is prone to severe privacy leakage as shown in ~\cite{our} and additionally requires heavy computation 
at each user node, which is in conflict with the FL objective of privacy preservation and conserving computation resources for 
users. 

\subsection{Evaluation of Privacy}

\begin{figure}[t]
\centering
\subfloat[]{\includegraphics[width=1.5in,height=1in]{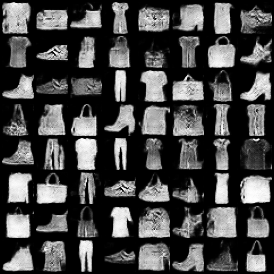}%
\label{fig2_1}}
\hfil
\subfloat[]{\includegraphics[width=1.5in,height=1in]{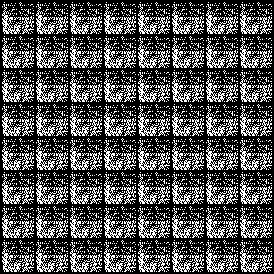}%
\label{fig2_2}}
\caption{Generated images from (a) cloud-server's model. (b) attacker's model.}
\label{fig_sim}
\end{figure}

We now evaluate the privacy leakage of the proposed UFed-GAN with respect to the attacker $\mathcal{A}$ as described in Section \ref{atk}.
Suppose that $\mathcal{A}$ has access to each communication round. 
We initialize the generator of $\mathcal{A}$ as $G_{\mathcal{A}}$, with some random weights and eavesdrop on user uplink and access $\mathcal{M}(\theta)$. $\mathcal{A}$ trains $G_{\mathcal{A}}$ similarly as with the server-side training. The design of  $G_{\mathcal{A}}$ is constrained by two parameters, the exact generator architecture and the initial weights $W_G$ of the generator at the server. Therefore, any attacker selecting the accurate generator architecture and initial weights is practically unachievable. 
However, in our experiments, we assume $\mathcal{A}$ knows the exact architecture of $G$, but with different random initial weights $W_{\mathcal{A}}\neq W_G$. We compare the generated images of the $G$ and $G_{\mathcal{A}}$ trained on the FashionMNIST dataset in Fig.~\ref{fig_sim}. It can be clearly seen that the generator $G$ on the cloud server captures the user's underlying data distribution, whereas $G_{\mathcal{A}}$ converges to a trivial point. Moreover, almost no useful visualization information is gained by the attacker as shown in Fig.~\ref{fig_sim}. The main reason is the uniqueness of the generator and the discriminator pair as presented in Proposition 1.

\begin{table}[!t]

\caption{FID score, IS score, and SSIM of the $\mathcal{A}$ and the cloud server after 100 communication rounds on the FashionMNIST dataset.} 
\label{tab:table1}
\centering
\begin{tabular}{|c||c||c|}
\hline
Metric & Attaker & Cloud Server\\
\hline
FID & 566.83 & 172.06\\ 
\hline
IS & 1.01 & 3.15\\ 
\hline
SSIM & 0.0067 & 0.8191\\ 
\hline
\end{tabular}
\end{table}

To examine privacy leakage quantitatively, we 
use the Frechet Inception Distance 
(FID) score~\cite{FID}, IS, and 
structural similarity index measure (SSIM)~\cite{realssim}. 
As discussed in the paper~\cite{ssim}, the similarity between the generated images and real data quantifies the privacy leakage. In table~\ref{tab:fid2}, we record average FID, IS, and SSIM scores for the data generated by $\mathcal{A}$ and the cloud server. Lower FID values, higher SSIM, and larger IS values represent better reconstruction quality. The FID score for $\mathcal{A}$ is higher than the cloud server by a great margin. This shows that, compared to the cloud server, $\mathcal{A}$ generated data carries less information about the training data. We further corroborate this observation by comparing the SSIM and IS values as well. The experimental results demonstrate the privacy preservation of UFed-GAN against full-GAN sharing.
\begin{table}[htb]
\label{tab:fid2}
\end{table}
\vspace{-1cm}
\section{Conclusion}
\label{sec:5}
In this work, we develop a novel framework of UFed-GAN to address the challenges imposed
by the lack of labeled data and by limited local computation resources in federated learning. Moreover, we propose a separate training strategy and a sharing scheme based on DCGANs. We provide an analysis of the convergence and privacy 
leakage of the proposed framework. Our empirical results demonstrate the superior performance of UFed-GAN 
in comparison against benchmark FL methods. We plan to investigate the communication overhead reduction for GAN-based FL and the application of semantic learning in distributed learning in future works.

\bibliographystyle{IEEEtran}
\bibliography{bib}

\begin{thebibliography}{10}
\providecommand{\url}[1]{#1}
\csname url@samestyle\endcsname
\providecommand{\newblock}{\relax}
\providecommand{\bibinfo}[2]{#2}
\providecommand{\BIBentrySTDinterwordspacing}{\spaceskip=0pt\relax}
\providecommand{\BIBentryALTinterwordstretchfactor}{4}
\providecommand{\BIBentryALTinterwordspacing}{\spaceskip=\fontdimen2\font plus
\BIBentryALTinterwordstretchfactor\fontdimen3\font minus
  \fontdimen4\font\relax}
\providecommand{\BIBforeignlanguage}[2]{{%
\expandafter\ifx\csname l@#1\endcsname\relax
\typeout{** WARNING: IEEEtran.bst: No hyphenation pattern has been}%
\typeout{** loaded for the language `#1'. Using the pattern for}%
\typeout{** the default language instead.}%
\else
\language=\csname l@#1\endcsname
\fi
#2}}
\providecommand{\BIBdecl}{\relax}
\BIBdecl

\bibitem{fedAvg}
B.~McMahan, E.~Moore, D.~Ramage, S.~Hampson, and B.~A.~y. Arcas,
  ``{Communication-efficient learning of deep networks from decentralized
  data},'' in \emph{Proceedings of the 20th International Conference on
  Artificial Intelligence and Statistics}, Lauderdale, FL, USA, Apr. 2017, pp.
  1273--1282.

\bibitem{non-iid1}
Y.~Zhao, M.~Li, L.~Lai, N.~Suda, D.~Civin, and V.~Chandra, ``Federated learning
  with non-iid data,'' \emph{arXiv preprint arXiv:1806.00582}, 2018.

\bibitem{fedprox}
T.~Li, A.~K. Sahu, M.~Sanjabi, M.~Zaheer, A.~Talwalkar, and V.~Smith, ``On the
  convergence of federated optimization in heterogeneous networks,''
  \emph{arXiv preprint arXiv:1812.06127}, 2018.

\bibitem{yuan2022convergence}
X.~Yuan and P.~Li, ``On convergence of fedprox: Local dissimilarity invariant
  bounds, non-smoothness and beyond,'' in \emph{Advances in Neural Information
  Processing Systems}, New Orleans, LA, USA, Jul. 2022, pp. 10\,752--10\,765.

\bibitem{GAN1}
I.~Goodfellow, J.~Pouget-Abadie, M.~Mirza, B.~Xu, D.~Warde-Farley, S.~Ozair,
  A.~Courville, and Y.~Bengio, ``Generative adversarial nets,'' in
  \emph{Advances in Neural Information Processing Systems}, vol.~27, Montreal,
  Canada, Dec. 2014.

\bibitem{cGAN1}
H.~Zhang, Z.~Zhang, A.~Odena, and H.~Lee, ``Consistency regularization for
  generative adversarial networks,'' in \emph{International Conference on
  Learning Representations}, Addis Ababa, Ethiopia, Apr. 2020.

\bibitem{fullGAN1}
X.~Cao, G.~Sun, H.~Yu, and M.~Guizani, ``Perfed-gan: personalized federated
  learning via generative adversarial networks,'' \emph{IEEE Internet of Things
  Journal}, vol.~10, no.~5, pp. 3749--3762, Mar. 2023.

\bibitem{our}
A.~Wijesinghe, S.~Zhang, and Z.~Ding, ``Ps-fedgan: an efficient federated
  learning framework based on partially shared generative adversarial networks
  for data privacy,'' \emph{arXiv preprint arXiv:2305.11437}, 2023.

\bibitem{mu2020computation}
J.~Mu, D.~Xie, H.~Huang, and X.~Jing, ``Computation-constrained spectrum
  sensing in iot-based scenarios,'' \emph{IET Communications}, vol.~14, no.~20,
  pp. 3631--3638, Nov. 2020.

\bibitem{learn_from_unl}
J.~Bekker and J.~Davis, ``Learning from positive and unlabeled data: a
  survey,'' \emph{Machine Learning}, vol. 109, pp. 719--760, Apr. 2020.

\bibitem{cluster2}
D.~K. Dennis, T.~Li, and V.~Smith, ``Heterogeneity for the win: One-shot
  federated clustering,'' in \emph{Proceedings of the 38th International
  Conference on Machine Learning}, Jul. 2021, pp. 2611--2620.

\bibitem{cluster1}
A.~Ghosh, J.~Chung, D.~Yin, and K.~Ramchandran, ``An efficient framework for
  clustered federated learning,'' in \emph{Advances in Neural Information
  Processing Systems}, Dec. 2020, pp. 19\,586--19\,597.

\bibitem{han2022fedx}
S.~Han, S.~Park, F.~Wu, S.~Kim, C.~Wu, X.~Xie, and M.~Cha, ``Fedx: Unsupervised
  federated learning with cross knowledge distillation,'' in \emph{European
  Conference on Computer Vision}, Tel Aviv, Israel, Nov. 2022, pp. 691--707.

\bibitem{9348203}
M.~Servetnyk, C.~C. Fung, and Z.~Han, ``Unsupervised federated learning for
  unbalanced data,'' in \emph{GLOBECOM 2020 - 2020 IEEE Global Communications
  Conference}, Taipei, Taiwan, Dec. 2020, pp. 1--6.

\bibitem{unlabeledFL}
N.~Lu, Z.~Wang, X.~Li, G.~Niu, Q.~Dou, and M.~Sugiyama, ``Federated learning
  from only unlabeled data with class-conditional-sharing clients,'' in
  \emph{International Conference on Learning Representations}, Apr. 2022.

\bibitem{orchestra}
E.~S. Lubana, C.~I. Tang, F.~Kawsar, R.~P. Dick, and A.~Mathur, ``Orchestra:
  unsupervised federated learning via globally consistent clustering,''
  \emph{arXiv preprint arXiv:2205.11506}, 2022.

\bibitem{dcGAN}
A.~Radford, L.~Metz, and S.~Chintala, ``Unsupervised representation learning
  with deep convolutional generative adversarial networks,'' \emph{arXiv
  preprint arXiv:1511.06434}, 2015.

\bibitem{IS}
T.~Salimans, I.~Goodfellow, W.~Zaremba, V.~Cheung, A.~Radford, and X.~Chen,
  ``Improved techniques for training gans,'' in \emph{Advances in neural
  information processing systems}, Barcelona, Spain, Dec. 2016.

\bibitem{moco}
K.~He, H.~Fan, Y.~Wu, S.~Xie, and R.~Girshick, ``Momentum contrast for
  unsupervised visual representation learning,'' in \emph{Proceedings of the
  IEEE/CVF conference on computer vision and pattern recognition}, Jun. 2020,
  pp. 9729--9738.

\bibitem{hayes2017logan}
J.~Hayes, L.~Melis, G.~Danezis, and E.~De~Cristofaro, ``Logan: evaluating
  privacy leakage of generative models using generative adversarial networks,''
  \emph{arXiv preprint arXiv:1705.07663}, pp. 506--519, 2017.

\bibitem{zhang2020federated}
F.~Zhang, K.~Kuang, Z.~You, T.~Shen, J.~Xiao, Y.~Zhang, C.~Wu, Y.~Zhuang, and
  X.~Li, ``Federated unsupervised representation learning,'' \emph{arXiv
  preprint arXiv:2010.08982}, 2020.

\bibitem{cifar10}
\BIBentryALTinterwordspacing
A.~Krizhevsky, ``Learning multiple layers of features from tiny images,'' pp.
  32--33, 2009. [Online]. Available:
  \url{https://www.cs.toronto.edu/~kriz/learning-features-2009-TR.pdf}
\BIBentrySTDinterwordspacing

\bibitem{svhn}
Y.~Netzer, T.~Wang, A.~Coates, A.~Bissacco, B.~Wu, and A.~Y. Ng, ``Reading
  digits in natural images with unsupervised feature learning,'' in \emph{NIPS
  Workshop on Deep Learning and Unsupervised Feature Learning 2011}, Granada,
  Spain, Dec. 2011.

\bibitem{fashionMnist}
H.~Xiao, K.~Rasul, and R.~Vollgraf, ``Fashion-mnist: a novel image dataset for
  benchmarking machine learning algorithms,'' \emph{arXiv preprint
  arXiv:1708.07747}, 2017.

\bibitem{chen2020simple}
T.~Chen, S.~Kornblith, M.~Norouzi, and G.~Hinton, ``A simple framework for
  contrastive learning of visual representations,'' in \emph{Proceedings of the
  37th International Conference on Machine Learning}, vol. 119, Vienna,
  Austria, Jul. 2020, pp. 1597--1607.

\bibitem{grill2020bootstrap}
J.-B. Grill, F.~Strub, F.~Altch{\'e}, C.~Tallec, P.~Richemond, E.~Buchatskaya,
  C.~Doersch, B.~Avila~Pires, Z.~Guo, M.~Gheshlaghi~Azar \emph{et~al.},
  ``Bootstrap your own latent-a new approach to self-supervised learning,''
  \emph{Advances in Neural Information Processing Systems}, vol.~33, pp.
  21\,271--21\,284, Dec. 2020.

\bibitem{li2020prototypical}
J.~Li, P.~Zhou, C.~Xiong, and S.~C. Hoi, ``Prototypical contrastive learning of
  unsupervised representations,'' \emph{arXiv preprint arXiv:2005.04966}, 2020.

\bibitem{zhuang2021collaborative}
W.~Zhuang, X.~Gan, Y.~Wen, S.~Zhang, and S.~Yi, ``Collaborative unsupervised
  visual representation learning from decentralized data,'' in
  \emph{Proceedings of the IEEE/CVF International Conference on Computer
  Vision}, Oct. 2021, pp. 4912--4921.

\bibitem{FID}
M.~Heusel, H.~Ramsauer, T.~Unterthiner, B.~Nessler, and S.~Hochreiter, ``Gans
  trained by a two time-scale update rule converge to a local nash
  equilibrium,'' in \emph{Advances in Neural Information Processing Systems},
  vol.~30, Long Beach, CA, USA, Dec. 2017.

\bibitem{realssim}
Z.~Wang, A.~Bovik, H.~Sheikh, and E.~Simoncelli, ``Image quality assessment:
  from error visibility to structural similarity,'' \emph{IEEE Transactions on
  Image Processing}, vol.~13, no.~4, pp. 600--612, 2004.

\bibitem{ssim}
D.~Dangwal, V.~T. Lee, H.~J. Kim, T.~Shen, M.~Cowan, R.~Shah, C.~Trippel,
  B.~Reagen, T.~Sherwood, V.~Balntas \emph{et~al.}, ``Analysis and mitigations
  of reverse engineering attacks on local feature descriptors,'' \emph{arXiv
  preprint arXiv:2105.03812}, 2021.

\end{thebibliography}

\newpage

 


\vspace{11pt}


\vfill

\end{document}